\documentclass[10pt,twocolumn,letterpaper]{article}
\usepackage{wacv}
\usepackage{times}
\usepackage{epsfig}
\usepackage{graphicx}
\usepackage{amsmath}
\usepackage{amssymb}
\usepackage{microtype} 
\usepackage{enumitem}
\usepackage{multirow}
\usepackage{floatrow}
\usepackage{comment}
\usepackage{float}
\usepackage{array}
\usepackage{amsthm}
\usepackage{algpseudocode}
\usepackage{algorithm}
\usepackage{algorithmicx}
\usepackage{booktabs}
\usepackage{footnote}
\usepackage[flushleft]{threeparttable}
\usepackage[accsupp]{axessibility}  % Improves PDF readability for those with disabilities.

\newfloatcommand{capbtabbox}{table}[][\FBwidth]
\newtheorem{Theorem}{Theorem}
\newtheorem{Lemma}{Lemma}

\newtheorem{innercustomthm}{Theorem}
\newenvironment{customthm}[1]
  {\renewcommand\theinnercustomthm{#1}\innercustomthm}
  {\endinnercustomthm}
  
\newtheorem{innercustomlemma}{Lemma}
\newenvironment{customlemma}[1]
  {\renewcommand\theinnercustomlemma{#1}\innercustomlemma}
  {\endinnercustomlemma}

\def\wacvPaperID{0614} % Enter the WACV Paper ID here

\wacvfinalcopy % *** Uncomment this line for the final submission

\ifwacvfinal
\fi
\ifwacvfinal
\usepackage[breaklinks=true,bookmarks=false]{hyperref}
\else
\usepackage[pagebackref=true,breaklinks=true,colorlinks,bookmarks=false]{hyperref}
\fi

\begin{document}

%%%%%%%%% TITLE
\title{Geometry-Aware Hierarchical Bayesian Learning on Manifolds}

\author{Yonghui Fan\\
Arizona State University\\
Tempe, Arizona\\
{\tt\small yfan61@asu.edu}
\and
Yalin Wang\\
Arizona State University\\
Tempe, Arizona\\
{\tt\small ylwang@asu.edu}
}

\maketitle
\ifwacvfinal
\thispagestyle{empty}
\fi

%%%%%%%%% ABSTRACT
\begin{abstract}
Bayesian learning with Gaussian processes demonstrates encouraging regression and classification performances in solving computer vision tasks. However, Bayesian methods on 3D manifold-valued vision data, such as meshes and point clouds, are seldom studied. One of the primary challenges is how to effectively and efficiently aggregate geometric features from the irregular inputs. In this paper, we propose a hierarchical Bayesian learning model to address this challenge. We initially introduce a kernel with the properties of geometry-awareness and intra-kernel convolution. This enables geometrically reasonable inferences on manifolds without using any specific hand-crafted feature descriptors. Then, we use a Gaussian process regression to organize the inputs and finally implement a hierarchical Bayesian network for the feature aggregation. Furthermore, we incorporate the feature learning of neural networks with the feature aggregation of Bayesian models to investigate the feasibility of jointly learning on manifolds. Experimental results not only show that our method outperforms existing Bayesian methods on manifolds but also demonstrate the prospect of coupling neural networks with Bayesian networks.
\end{abstract}

%%%%%%%%% BODY TEXT
\section{Introduction}

\begin{figure}[t]
\centering
\includegraphics[width=\linewidth]{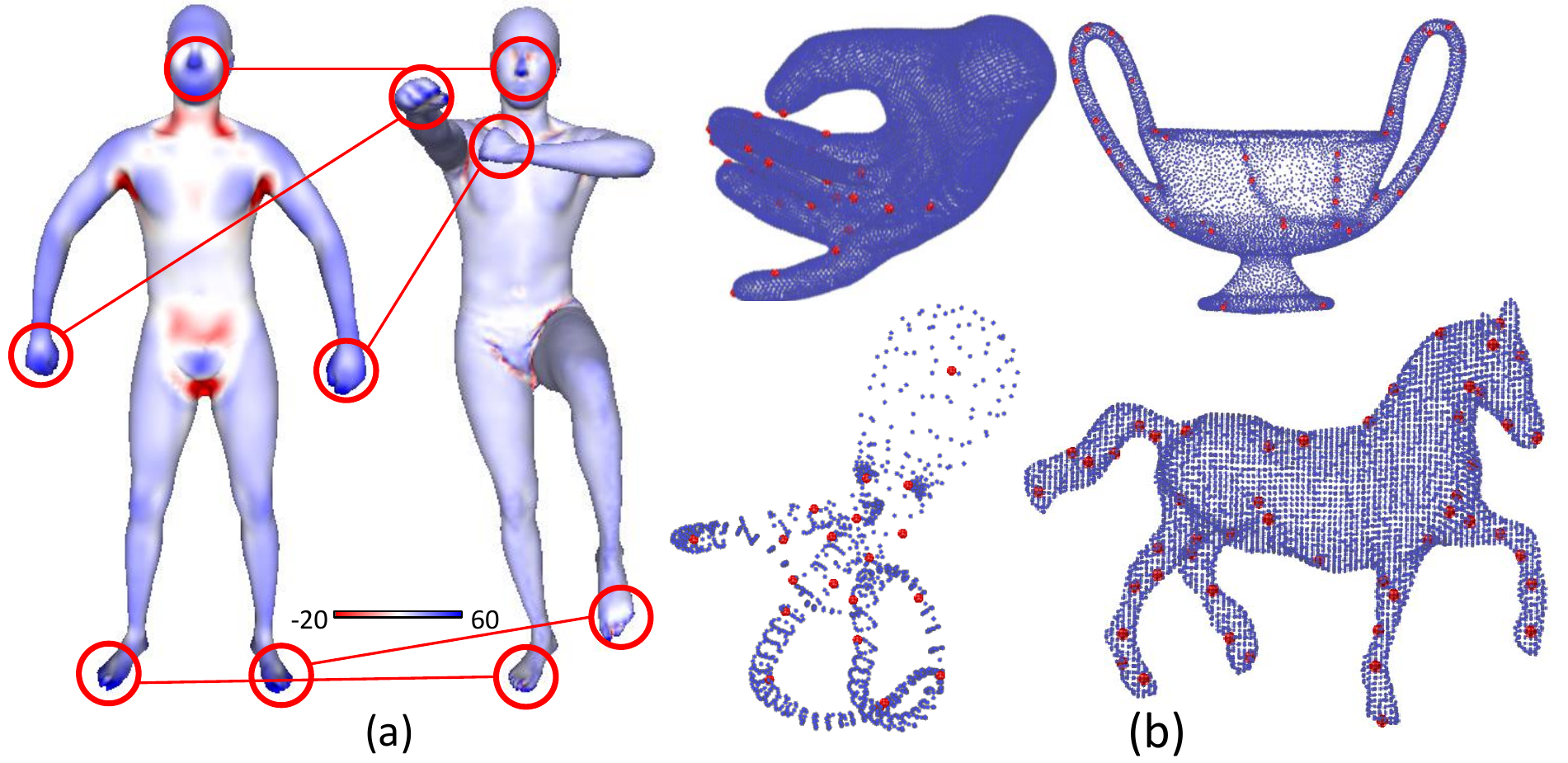}
\caption{Examples of manifold-valued data. (a) Triangle meshes of two human poses. Meshes are rendered by normalized mean curvatures by MeshLab~\cite{LocalChapterEvents:ItalChap:ItalianChapConf2008:129-136}. ROIs are marked by red circles; (b) Point clouds of different objects from McGill 3D shape benchmark~\cite{siddiqi2008retrieving}. Salient points based on GP regression are marked by red spheres.}
\label{fig:datatype}
\end{figure}

Three-dimensional data on Riemannian manifolds, such as triangle meshes and point clouds as shown in Figure~\ref{fig:datatype}, is widely used to describe the shape information in object understanding, scene understanding, and many other vision tasks. Extracting and aggregating geometric features is considered the key to leveraging the intrinsic shape information of this type of data~\cite{bronstein2017geometric}. %Many closed-form geometric feature descriptors~\cite{aubry2011wave,bronstein2010scale,fanhfs} or learning-based neural network methods~\cite{yin2016multichannel,qi2017pointnet,feng2019hypergraph} can provide expressive description of the geometric property, but how to effectively and efficiently aggregate features becomes the next challenge because the manifold-valued data usually lacks organized on-the-grid data structures and uniform data sizes~\cite{hua2018pointwise, gao2019gaussianearly,fan2020morphometric}.
% Deep neural network~(NN) based learning methods represent the current state-of-the-art approaches~\cite{litany2017deep,sung2018deep,feng2018gvcnn,you2019pvrnet}. %, such as the deep functional map for dense shape correspondence, and pointnet++~\cite{qi2017pointnet++} for point cloud segmentation and classification ect.%, for example using multi-layer perceptrons~(MLP) or fully convolutional networks~(FCN)~\cite{sung2018deep,feng2018gvcnn,you2019pvrnet}. 
Recently, the Gaussian process~(GP) based Bayesian learning emerges to be a study hotspot%achieve impressive performances in many regression and classification tasks on images
~\cite{van2017convolutional,blomqvist2019deep,fan2020convolutional}. Theoretically, it has been proven that a single fully connected neural network~(NN) layer with an infinity width is essentially a GP~\cite{neal1996priors}. Further, this equivalence is extended to deep fully connected NNs and hierarchically connected GPs~\cite{garriga2019deep,lee2018deep}. Practically, encouraging results have been demonstrated in various applications~\cite{borovitskiy2020mat}. %Bayesian  in propagating and integrating features~\cite{lee2018deep,dutordoir2020bayesian}. %Although current vision applications of Bayesian learning techniques are mainly based on simple image datasets, 
In this work, we focus on developing GP-based Bayesian learning methods for solving vision tasks on manifolds. Specifically, we concentrate on two fundamental aspects: GP kernel design and Bayesian learning framework on manifolds.

Since the property of a zero-mean GP is largely determined by its kernel function~\cite{rasmussen2004gaussian}, a primary goal is to develop an expressive kernel. Essentially, we aim at integrating two important characteristics into a shallow kernel structure: \textit{geometry-awareness} and \textit{intra-kernel convolution}.

\textit{Geometry-awareness} stresses the capability of learning geometric features in the prior knowledge so that the posterior inference respects the representative regions of a 3D shape~\cite{gao2019gaussian}. For example, when distinguishing different human poses, red circled regions of interest (ROIs) in Figure~\ref{fig:datatype}(a) are expected to be numerically highlighted because their regional features are more geometrically significant. The current strategy of achieving geometry-awareness is to directly add geometric feature descriptors to the kernel design~\cite{gao2019gaussianearly,fan2020morphometric}. However, this strategy heavily relies on computing specific hand-crafted features, which potentially impedes the generality to broader types of applications. We propose a paradigm shift where only the point coordinates are needed. It enables our kernel to be geometry-aware on all commonly-used types of manifold-valued data. 

\textit{Intra-kernel convolution} introduces the convolutional filtering to the kernel construction so that the GP inference has a powerful feature aggregation ability~\cite{van2017convolutional}. %Being motivated by the success of convolutional NN~(CNN), the 
This characteristic has been widely studied to increase the expressiveness of GPs~\cite{walker2019graph,dutordoir2020bayesian}. Most of the previous work uses the additive patch-wised computational structure to explicitly mimic the mechanism of convolutional NNs~(CNNs)~\cite{durrande2012additive, van2017convolutional}. But this approach is not feasible for manifold-valued data because of its off-the-grid structure. As one attempt, the graph convolutional GP~(GCGP) in~\cite{walker2019graph} used local coordinates to adjust the inputs to a uniform on-the-grid style. However, some drawbacks, such as the huge computational cost and the strict requirement on the input size, are noticed. Alternatively, we propose an implicit intra-kernel convolution. Mathematically, we rigorously show that the convolutional filtering can be delicately embedded into the kernel definition.%Mathematically, it shows that the convolutional filtering it is possible to implicitly compact both properties into one kernel.  

In this paper, we propose a hierarchical Bayesian model for manifold-valued tasks. The core is a kernel derived from a stochastic partial differential equation~(SPDE) that generalizes a real physical process called \textit{periodic potential diffusion process}. Two observations explaining why we choose this particular physical process are discussed in Sec.~\ref{sec:kernel}. Mathematically, we prove that the kernel implicitly integrates both the mean curvature flow, which is an effective geometric feature descriptor in $\mathcal{R}^3$, and a convolutional filtering. %A GP layer is defined by the new kernel and our Bayesian network is concatenated with multiple GP layers. 
For tackling the irregular input dimensionality, we firstly use a GP-based salient point selection algorithm to obtain a uniform and light input, then, feeding it to the Bayesian network. Additionally, because the input of a Bayesian model on manifolds is the geometric features and NNs are strong in learning expressive features, we explore the potential of incorporating NNs with hierarchical Bayesian methods to leverage the strengths of both methods. Our contributions are summarized into three-folds:
\newline(1) A kernel with both geometry-awareness and intra-kernel convolution properties. No hand-crafted feature is needed for involving geometric properties. The method is feasible to all commonly-used manifold-valued data;
\newline(2) A Bayesian network for manifold-valued tasks, including a salient point selection module that non-linearly reduces the data dimensionality and organizes the irregular inputs; 
\newline(3) An exploration on NN+Bayesian approaches to leverage both the feature learning ability of NNs and the feature aggregation ability of the Bayesian methods.

Both empirical and numerical experimental results verify the effectiveness of our methods. We hope this work not only makes contributions to the Bayesian learning methods on manifolds but also sheds new light on integrating different learning mechanisms to maximize their learning power.

\section{Related Work}
Kernel design is always an important topic in GP-based Bayesian learning studies. Our initial thought originated from the problem addressed by Stein in~\cite{stein2012interpolation} that the infinite differentiability of a Gaussian kernel led to an unrealistic match with the physical processes. Later, Stein proposed the well-known Mat\'ern kernel family as a generalization of the Gaussian radial basis functions~(RBF) to solve this problem. An intriguing idea is: why not derive a kernel function directly from the expression of a physical process? In this way, the kernel will intrinsically come with a real physical rationale. This idea was further strengthened by S{\"a}rkk{\"a}'s statement~\cite{sarkka2011linear,solin2020hilbert} that any SPDE was a potential kernel. Given the fact that many physical processes are generalized by SPDEs, it is intuitive to combine the above two opinions together as a trustful theoretical foundation of kernel development~\cite{lindgren2011explicit}. For example, the Mat\'ern kernel is actually the solution of a linear fractional SPDE~\cite{cressie2015statistics,sherman2011spatial,guttorp2006studies}.%: $(\kappa^2-\Delta )^{\alpha/2}X(v)=W(v),\quad v\in \mathbb{R}^d$, where $X(v)$ is a Gaussian field function about spatial location $v$, $W(v)$ is the Gaussian white noise, $\alpha$ is often assumed to be a constant, and $\kappa$ is the parameter.

However, classical kernels mainly dealt with data in the Euclidean space and considered less on the Riemannian manifolds. One solution was using Riemannian metric and space mapping techniques to achieve the domain transform~\cite{fahrmeir2013regression,mallasto2018wrapped}. %The central idea of these methods is to transform the metric space to pursue a reasonable spatial measurement.% instead of harnessing the geometric feature space. Additionally, 
But clearly, this approach was not feasible to the data like volumetric meshes and point clouds. Alternatively, wrapping geometric features into kernels was proven to be effective~\cite{prisacariu2011nonlinear,castillo2014thomas,lin2018extrinsic,gao2019gaussian,fan2020morphometric}. For example, the weighted GP~(W-GP)~\cite{gao2019gaussianearly} yielded reasonable inferences after weighing the RBF kernel with the mean curvature and Gaussian curvature; the morphometric GP~\cite{fan2020morphometric} used wave kernel signature metric and demonstrated good performances on 3-dimensional manifolds. %Other applications, such as the intrinsic framework in~\cite{castillo2014thomas} and the extrinsic framework in~\cite{lin2018extrinsic}, also demonstrate inspiring results after involving proper geometric features. 
These methods used explicit geometric expressions, which often relied on specific simplicial complex. Conversely, we implement an implicit geometric expression which is only sensitive to the distance lag.

Bayesian learning architecture is also an emerging topic. Additive GP~\cite{durrande2012additive} directly enabled the convolutional GP~(CGP)~\cite{van2017convolutional}. We also use additive structure in our kernel design. The aforementioned GCGP is an implementation of CGP on graphs. It adjusted the irregular inputs by referring to the method used in graph convolutional networks~(GCNs). Instead, we use a manifold learning strategy to organize the inputs. The studies of sparse variational GP and posterior estimation facilitated the development of deep GP~(DGP)~\cite{damianou2013deep, sheth2015sparse,hensman2015mcmc, blomqvist2019deep}. In our hierarchical Bayesian model, we follow the framework of DGPs and use the doubly stochastic variational inference method~\cite{salimbeni2017doubly}.

\section{Preliminaries}
Some notations are defined here. Given a manifold-valued data $\mathcal{M}=(\mathcal{V}, \mathcal{E},\mathcal{F})\in\mathbb{R}^d$ with a vertex set $\mathcal{V}$ of size $|\mathcal{V}|$, an edge set $\mathcal{E}$ of size $|\mathcal{E}|$ and a face set $\mathcal{F}$ of size $|\mathcal{F}|$. $\mathcal{E}$ and $\mathcal{F}$ can be empty. A distance lag between vertex $v\in\mathcal{V}$ and its neighborhood $v'$ is denoted as $\left \| v \right \|=\left \| v-v' \right \|^2$. Define a GP, ${GP}(m, K)$, as a random process where the joint distribution of a finite collection of observations $Y=\{y_1,...,y_n\}$ of samples $X=\{x_1,...,x_n\}$ follows a multivariate Gaussian distribution: $p(Y|X)\sim\mathcal{N}(m, K)$, where $m$ is the mean function and $K$ is the covariance function or kernel. The dimension of the kernel matrix $K$ is denoted by subscripts. %Some other symbols are defined later.
\subsection{Periodic Potential Diffusion Process}\label{sec:3.1}
Our kernel derivation originates from the periodic potential diffusion process. It is a special case of the reaction diffusion process. A reaction diffusion process $T(v,t)$ is the solution to a reaction diffusion equation~\cite{ehrlich1980surface,strauss2013partielle}:
\begin{equation} \label{eq:spde}
\vspace{-0.5em}
\frac{\partial T(v,t)}{\partial t} = \alpha\Delta T(v,t)+F(v,t), \quad t\geq 0
\vspace{-0.2em}
\end{equation}
% \begin{equation}\label{eq:2}
% \vspace{-0.2em}
% T(v,t)=e^{-t\Delta}f+\int_{0}^{t}e^{-(t-s)\Delta}F(s)ds
% \end{equation}
where $\Delta$ is the Laplace operator, constant $\alpha$ is 1. The initial condition is 0. $F$ is the reaction function that defines the property of the energy source~\cite{ecker2008heat}. Given a certain $F$, there exists a corresponding physical scenario~\cite{strauss2013partielle,oksendal2013stochastic}. 
%For example, when $F=0$, it is the heat equation. %Eq.~\eqref{eq:spde} is a general expression of many physical equations that can be expressed as a second order SPDE,  %For example, in a heat diffusion case, when $F$ defines properties of the heat source, the reaction diffusion process is a temperature distribution that records the temperature at a certain time and location. When $F=0$, Eq.~\ref{eq:spde} is the classical heat equation. Eq.~\eqref{eq:spde} is a general expression of many physical equations that can be expressed as a second order SPDE.
The reaction function $F$ can be expressed as the multiplication of a Dirac delta function at location $v$ and a temporal function $h(t)$: $F=h(t)\delta(v- v')$. By defining the Green's function of Laplace operator under the Dirichlet boundary condition as $G(v, v',t)$~\cite{beck1992heat}, $T$ is equal to:
\begin{equation}\label{eq:2}
\vspace{-0.2em}
T(v, v',t)=\int_{0}^{t}G(v, v',t-s)h(s)ds
\vspace{-0.2em}
\end{equation}

%Noting that Eq.~\eqref{eq:2} is equal to Eq.~\eqref{eq:3}. 
Reminding that the Green's function in an $\mathbb{R}^d$ diffusion problem has the standard form: $G=\frac{e^{-v^2/4 t}}{(4\pi t)^{d/2}}$. When $h(t)$ is periodic: $h(t)=cos(\omega t)$, Eq.~\eqref{eq:2} is further derived as:
\begin{equation}\label{eq:3}
\vspace{-0.2em}
T=\int_{0}^{t}cos(\omega(t-s))\frac{e^{-v^2/4 s}}{(4\pi t)^{d/2}}ds
\vspace{-0.2em}
\end{equation}
Eq.~\eqref{eq:3} is called the \textit{periodic potential diffusion process}, which is the theoretical foundation of our kernel.
%\noindent\textbf{Notations:} Given a manifold-valued data $\mathcal{G}=\{\mathcal{V}, \mathcal{E}, \mathcal{F}\}$ in $\mathbb{R}^3$, where $\mathcal{V}$ is the indexed vertex set containing totally $|\mathcal{V}|$ vertices. $\mathcal{E},\mathcal{F}$ and $\mathcal{T}$ are indexed edge set, and face set. $\mathcal{E} and \mathcal{F}$ can be empty in some types of data. $v_n$ is the $n^{th}$ vertex. %$\tilde{v}^n$ is the $n^{th}$ salient vertex. Notice that we also use $v$ to stand for an arbitrary spatial point/variable on manifold for theory analysis. We use $t$ to stand for a temporal variable. 
%Define a GP as $\mathcal{GP}(m,K)$, where $m$ is a mean function and $K$ is the kernel function. Here, we interchangeably use kernel function, kernel and covariance function to represent any positive semi-definite matrix. In this paper, all kernels are stationary. Using $\left \| \cdot  \right \|$ to denote the distance lag. For spatial variables, the distance can be L1 norm, L2 norm or some other metrics. For temporal variables, the distance is L1 norm. For reading convenience, some notations are defined near to the places they are used.

\subsection{Gaussian Process Regression}\label{sec:3.2}
%A GP, $\mathcal{GP}(m, K)$, is defined by a mean function $m$ and a kernel $K$. Here, we only consider zero mean cases. 
A GP regression~(GPR) aims at learning a multivariate distribution that fits with the training data and predicts the observation $y_{n+1}$ when a testing sample $x_{n+1}$ arrives~\cite{rasmussen2004gaussian}. The Bayes' theorem is used to transform the prior knowledge to posterior inference in the learning process. As known, every finite marginal distribution of a GP still follows a multivariate Gaussian distribution. Therefore, the predictive distribution $\mathcal{N}(m', K')$ can be uniquely determined by the standard rules for conditioning Gaussian distributions:
% \begin{equation}\label{Eq:5}
% \begin{pmatrix}
% Y_n\\ 
% Y_{n+1}
% \end{pmatrix}\sim \mathcal{GP}\begin{pmatrix}
% \begin{pmatrix}
% \mathbf{0}\\ 
% \mathbf{0}
% \end{pmatrix}, 
% \begin{pmatrix}
% K_n & K_{n,n+1}\\K_{n,n+1}^T 
% & K_{n+1}
% \end{pmatrix}
% \end{pmatrix}
% \end{equation}
\begin{equation}\label{Eq:4}
m'_{(n+1)\times1}=K_{n\times(n+1)}^T K_{n\times n}^{-1}Y_{n\times 1}
\end{equation}
\vspace{-.5em}
\begin{equation}\label{Eq:5}
    K'_{(n+1)\times (n+1)}=K_{(n+1)\times(n+1)}-K^T_{n\times (n+1)}K^{-1}_{n\times n}K_{n\times(n+1)}
\end{equation}
\vspace{-1em}
\begin{equation}\label{Eq:6}
\begin{split}
\begin{aligned}
&K_{n\times(n+1)} = K_{(n+1)\times n}^T=\\&\begin{pmatrix}
K(x_1,x_1) & \cdots  & K(x_1,x_n), & K(x_1,x_{n+1})\\ 
\vdots  &  &  & \vdots \\ 
K(x_n,x_1) & \cdots & K(x_n,x_n), & K(x_n,x_{n+1})
\end{pmatrix}
\end{aligned}
\end{split}
\end{equation}
\begin{equation}\label{Eq:7}
\vspace{-.5em}
\begin{split}
\begin{aligned}
&K_{(n+1)\times (n+1)}=\\ &\begin{pmatrix}
K(x_1,x_1) & \cdots  & K(x_1,x_n), & K(x_1,x_{n+1})\\ 
\vdots  &  &  & \vdots \\ 
K(x_n,x_1) & \cdots & K(x_n,x_n), & K(x_n,x_{n+1})\\
K(x_{n+1},x_1)& \cdots& K(x_{n+1},x_n), & K(x_{n+1},x_{n+1})
\end{pmatrix}
\end{aligned}
\end{split}
\vspace{-.5em}
\end{equation}
When new samples are continuously given, the GP is recursively updated. Later, we adopt the framework of GPR in a salient point selection algorithm. Each salient point is taken as a sample. We update the saliency map after adding the previous selection into the prior and then select the next one until a certain number of salient points are collected.
%\textbf{Motivations of choosing GP regression: }
%Despite the commonly-mentioned advantages of GP regression and Bayesian inference in spatial data analysis, we summarize two reasons of choosing GPR as the framework in the saliency extraction problem: (1) for implementation on large-scale data, GP regression is capable of working with active learning strategy~\cite{liang2015landmarking,gao2019gaussianearly}. At each time step, only one vertex is selected and one dimension is added to the kernel matrix after the current extraction. This consideration has a significant acceleration effect in computational efficiency; (2) GP is the best tool we can find to construct a process that allows for the geometric dependence varying as a function of distance. A stationary kernel function is a perfect platform for defining local spatial features. Only zero-mean GP is considered in this paper.
\vspace{-0.2em}
\subsection{Deep Gaussian Processes with Doubly Stochastic Variational Inference}\label{sec:HBM}
A DGP is a deep belief network that hierarchically concatenates multiple Gaussian process latent variable models together~(GP-LVMs)~\cite{damianou2013deep}. It mimics the composition of restricted Boltzmann machines~(RBMs) in NNs. %In a single GP layer, a GP prior is built to model all the values of a stochastic function $F$ as a joint Gaussian distribution $\mathcal{G}(m, K)$ given $N$ sample-observation pairs. The posterior inference is computed by the GPR. The original GPR has a cubic time complexity because of the inverse matrix computation. 
The sparse variational inference is usually used in GPR to estimate the posterior and avoid the cubic complexity~\cite{sheth2015sparse}. Suppose $M$ inducing points $Z=\{z_1,...,z_M\}(M\ll N)$ are selected, the complexity is decreased to $\mathcal{O}(M^2N)$ in a single GPR. For a DGP, the doubly stochastic variational inference is often applied to estimate the posterior~\cite{salimbeni2017doubly, blomqvist2019deep}. Specifically, the sparse variational inference is used to simplify the correlations within layers and keep the correlations between layers unchanged. In a DGP with $L$ layers, the prior is recursively defined on a series of vector-valued stochastic functions $F=\{F^1,..., F^L\}$. The $i^{th}$ row of $F^l$ is denoted as $f_i^l$. Function values at inducing points $Z$ are $U$. Each single function has an independent Gaussian prior and inducing points. A joint density of a DGP can be expressed as:
\begin{equation}\label{eq:dgp_joint_distribution}
\vspace{-.6em}
\begin{split}
&p(Y,\{F^l, U^l\}_{l=1}^L)=\\ &\underset{likelihood}{\underbrace{\prod_{i=1}^{N}p(y_i|f_i^L)}}\underset{prior}{\underbrace{\prod_{l=1}^{L}p(F^l|U^l;F^{l-1},Z^{l-1})p(U^l;Z^{l-1})}}
\end{split}
\vspace{-.7em}
\end{equation}
According to the theories of variational inference, a factorized form of the posterior joint density is defined as~\cite{salimbeni2017doubly}:
\begin{equation}\label{eq:variationaljointdensity}
\vspace{-0.5em}
\begin{split}
q(\{f^l, U^l\}^L_{l=1})&=\prod_{l=1}^{L}p(f^l|f^{l-1},U^l, Z^l)q(U^l)\\   
\end{split}
\end{equation}% 
where $q(U^l)$ is a Gaussian with mean function $m^l$ and covariance function $S^l$ for layer $l$. Eq.~\eqref{eq:variationaljointdensity} indicates that the prediction of the $l^{th}$ layer, $f^l$, depends on the previous prediction $f^{l-1}$ and the inducing points of the current layer. By marginalising the approximation $q(U^l)$ from each layer, the $i^{th}$ factorized variational posterior of the final layer is the integral of all paths $(f_i^1,...,f_i^L)$ through the Gaussian distributions defined by parameters $m^l$, and $S^l$:
\begin{equation}\label{eq:10}
\vspace{-0.5em}
q(f_i^L)=\int \prod_{l=1}^{L-1}q(f_i^l|m^l,S^l;f_i^{l-1},Z^{l-1})df_i^l
% \vspace{-0.5em}
\end{equation}
The objective function is the doubly stochastic evidence lower bound~(ELBO):
\begin{equation}\label{eq:elbo}
\vspace{-.5em}
\begin{split}
\mathcal{L}=\sum_{i=1}^{N}\mathbb{E}_{q(f_i^L)}[logp(y_i|f_i^L)]-\sum_{l=1}^{L}\mathcal{KL}(q(U^l)||p(U^l))
\end{split}
\vspace{-.7em}
\end{equation}
where $\mathcal{KL}$ is the Kullback–Leibler divergence. The ELBO has the complexity $\mathcal{O}(M^2N(D^1 +...+D^L))$ to compute, $D^l$ is the size of the $l^{th}$ layer. The variational expection likelihood $\mathbb{E}$ in Eq.~\ref{eq:elbo} is computed using the Monte Carlo approximation. 
%In a Bayesian model training process, the inversed ELBO is often used as the loss function.
Please refer to~\cite{salimbeni2017doubly} for more details.

\section{Methods}
\subsection{Geometry-Aware Convolutional Kernel}\label{sec:kernel}
In this section, we start by briefly explaining what motivates us to choose the periodic potential diffusion process as the theoretical basis. Then, we provide two implementations of the kernel for different applications. Furthermore, we introduce our theoretical analysis about the property of the kernel. In the end, a hierarchical Bayesian model is defined. 

The idea of using periodic potential diffusion process comes from two observations:
\\
\noindent\textbf{The first observation} is that the integral Laplace transform of function $f(t)=t^{d-1}e^{-\frac{1}{4}at}$ in $\mathbb{R}^d$ in Eq.~\eqref{eq:12}~(which is a special upper incomplete gamma function, $a$ is a constant)  has several similar terms with the Mat\'ern kernel in Eq.~\eqref{eq:13}:
\begin{equation}\label{eq:12}
\int_{0}^{\infty }t^{d-1}e^{-\frac{1}{4}at}e^{-st}dt=2\left [ (\frac{1}{4}a)^{\frac{1}{2}}s^{-\frac{1}{2}} \right ]^d \mathcal{K}_d(a^{\frac{1}{2}}s^{\frac{1}{2}})
\end{equation}
\begin{equation}\label{eq:13}
C(\tau )=\frac{\sigma^2}{\Gamma (d)2^{d-1}}(2\sqrt d \tau \kappa )^d \mathcal{K}_d(2\sqrt d \tau \kappa )
\end{equation}
Both equations have modified Bessel function of the second kind $\mathcal{K}_d$, and the rest parts are functions with the same dimension order $d$. This indicates the possibility of deriving a kernel from Eq.~\eqref{eq:12}.

\noindent\textbf{The second observation} is that the Green's function of Laplace-Beltrami operator $\Delta$ in the 3D diffusion problem belongs to the family of $t^{d-1}e^{-\frac{1}{4}at},d=3$. Bochner's theorem states that a stationary kernel $K$ is positive definite in $\mathbb{R}^d$ if it is the Fourier transform of a positive bounded measure function~\cite{bochner2005harmonic}. %$F$: $K(x,x')=\int _{\mathbb{R}^d}cos(\omega(x-x'))Fd\omega$. 
Taking Eq.~\eqref{eq:3} as the real part of the Fourier transform (decomposing the exponential term with Euler's formula), it is already a stationary kernel function regarding the temporal variable $t$. If the spatial part is also proven to be positive semi-definite~(PSD), then the periodic potential diffusion process $T$ is a valid kernel function.
	
Summarizing these two observations and incorporating the background in Sec.~\ref{sec:3.1}, our goal is to derive a close-form expression from Eq.~\eqref{eq:3} and prove that this expression is PSD regarding its spatial variable $v$. Unfortunately, the integral in Eq.~\eqref{eq:3} has no explicit solution according to~\cite{sarkka2011linear}. 

Alternatively, we apply the same approximation strategy in~\cite{fan2020convolutional} to estimate the Eq.~\eqref{eq:3} as the combination of a cosine Fourier transform$\hat{f_c}(\omega)$ and a sine Fourier transform$\hat{f_s}(\omega)$:% by assuming $t$ is $\infty$:
\begin{equation}\label{eq:14}
\begin{aligned}
%\begin{split}
&T=\\&cos(\omega t)\int_{0}^{t}cos (\omega s)G(s)ds+sin(\omega t)\int_{0}^{t}sin(\omega s)G(s)ds \\
&\approx cos(\omega t)\hat{f_c}(\omega)+sin(\omega t)\hat{f_s}(\omega)
%\end{split}
\end{aligned}
\end{equation} 
By solving this approximated form, we have the closed-form periodic potential diffusion process:
\begin{equation}\label{eq:21}
T=\frac{1}{4\pi}e^{-\left \| v \right \|\sqrt{\frac{1}{2}\omega}}cos(\left \| v \right \|\sqrt{\frac{1}{2}\omega}+\omega t)
\end{equation}
For simplicity, we define a frequency term $\lambda=\sqrt{\frac{1}{2}\omega}$ and a phase term $\phi=\omega t$. $\lambda$ and $\omega$ are hyper-parameters that are determined by regular parameter tuning methods. Reminding that the diffusion process is dynamic, we can express it as the accumulation of values at $N$ time slots. Therefore, the final kernel definition is expressed as an additive kernel~\cite{durrande2012additive}:
\begin{equation}\label{eq:kerneldefinition}
K(\left \| v \right \|, \lambda, \phi)=\frac{1}{4\pi}\sum_{n=1}^N e^{-\lambda_n\left \| v \right \|}cos(\lambda_n\left \| v \right \|+\phi_n)
\end{equation}
For regression tasks, the implementation in Eq.~\eqref{eq:kerneldefinition} is sufficient. But for better fitting with a hierarchical Bayesian model, we further introduce an implementation by taking the kernel as an ARD covariance function~\cite{gardner2018gpytorch}. An individual length-scale parameter $\alpha$ is added for each input
dimension which determines the relevancy of the input to the task:
\begin{equation}\label{eq:ardpdk}
%\vspace{-1em}
K(\left \| v \right \|, \lambda, \phi, \alpha)=\frac{1}{4\pi}\sum_{n=1}^N e^{-\lambda_n\frac{\left \| v \right \|}{\alpha_n}}cos(\lambda_n\frac{\left \| v \right \|}{\alpha_n}+\phi_n)
% \vspace{-0.5em}
\end{equation}
GPs defined with Eq.~\eqref{eq:kerneldefinition} or Eq.~\eqref{eq:ardpdk} can be taken as the mixture of GPs. Previous studies show that the mixture of Gaussian has a universal approximation ability in fitting with continuous distributions~\cite{tobar2015learning,parra2017spectral}.    

The next goal is to prove that Eq.~\eqref{eq:kerneldefinition} is PSD regarding its spatial variable. Eq.~\eqref{eq:kerneldefinition} is the composition of two functions: $e^{-\lambda\left \| v \right \|}$ and $cos(\lambda\left \| v \right \|+\phi)$. It is acknowledged that the exponential function and the cosine function are PSD regarding variable $v$. According to the composition property of a PSD function, the result of positive real function/constant times a PSD function is still PSD~\cite{bhatia2007positive}. So, Eq.~\eqref{eq:kerneldefinition} is spatially PSD. Eq.~\eqref{eq:kerneldefinition} can be taken as a compositional kernel.
\begin{figure}[t]
\centering
\includegraphics[width=\linewidth]{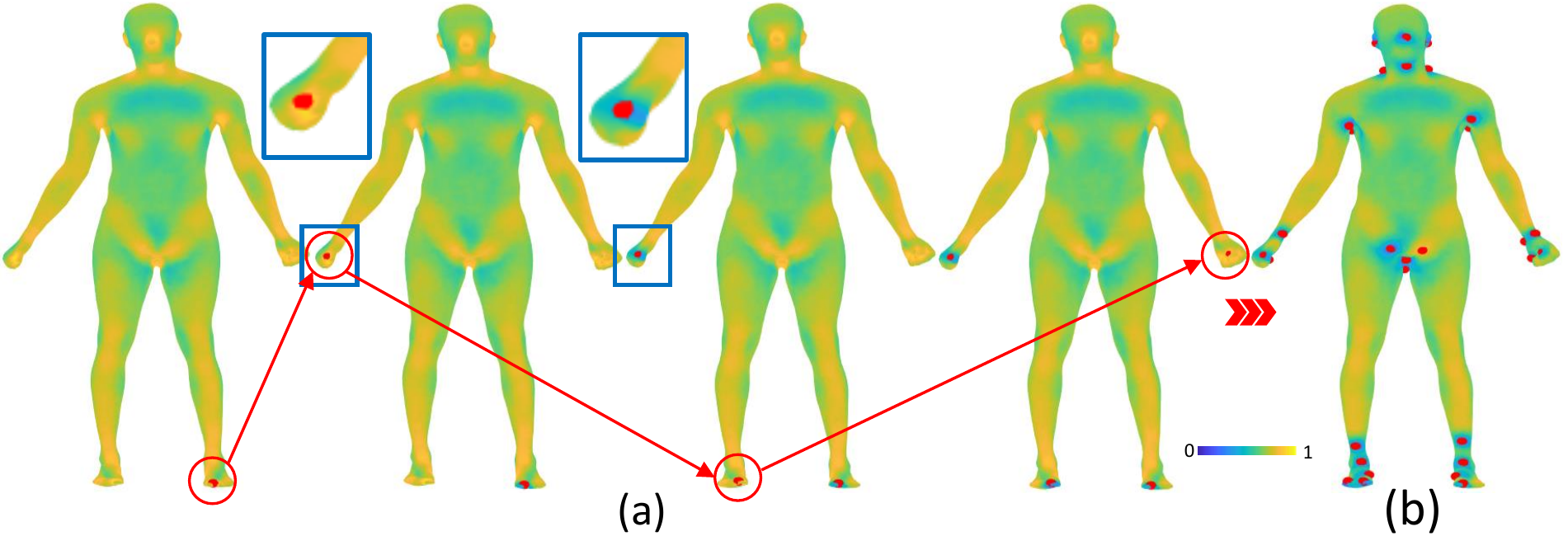}%21
\vspace{-0.5em}
\caption{Demonstration of salient point selection on a human pose model. The meshes are rendered by normalized uncertainty scores at each iteration. (a) The progress of selecting the first four salient points. Salient points are remarked by red spheres. Noticing the color change before and after the selection in two zoom-in regions. (b) The first $30$ salient points. The high uncertainty score regions are still centralized in ROIs after selecting the $30^{th}$ salient point.}
\label{fig:saliency}
\end{figure}
Eq.~\eqref{eq:spde} indicates a family of kernels. This opinion is generalized as:% It is said that a kernel based on a rational physics process guarantees the inference well matches with the inherent structures.
\begin{Theorem}\label{theorem:1}
A real-valued function $T(v,t)$ on $\mathbb{R}^d$ is a spatial-temporal kernel function if it is a linear/non-linear diffusion process: $\frac{\partial T}{\partial t}=\alpha \Delta T + P(t)\delta(v)$, where $\alpha$ is a positive constant, $P(t)$ is a periodic function, $\delta(v)$ is the Dirac delta function, and $\Delta$ is the Laplace operator.
\end{Theorem}

\begin{proof}
The proof is provided in Supplementary.
\end{proof}

% S{\"a}rkk{\"a} addressed that any stochastic partial differential equation has a potential to be a kernel~\cite{sarkka2011linear}. Theorem~\ref{theorem:1} provides an applicable class of solutions. Comparing with methods derived from analogous theory background, e.g. the heat kernel~(HK)~\cite{berline2003heat}, ours is in a dynamic process scenario, hence it allows more variations and non-linearity in matching with the data.
The kernel implemented in Eq.~\eqref{eq:kerneldefinition} or Eq.~\eqref{eq:ardpdk} is named as the geometry-aware convolutional~(GAC) kernel. The GAC kernel satisfies the two characteristics discussed in Introduction, which are summarized as two lemmas:
\begin{Lemma}\label{lemma:1}
The GAC Kernel embeds the mean curvature flow in $\mathbb{R}^3$, which enables it to be geometry-aware.
\end{Lemma}
% Lemma~\ref{lemma:2} explains why our kernel is geometry-aware.
\begin{Lemma}\label{lemma:2}
The GAC Kernel embeds a convolution filtering within the kernel structure, called intra-kernel convolution.
\end{Lemma}
% Their proofs are available in the Supplementary.
\begin{proof}
The proofs are provided in the Supplementary.
\end{proof}
We validate the proposed GAC kernel in two different studies. The first one is to use it in a regular GPR model. An unsupervised salient point selection algorithm will be introduced. This implementation fully utilizes geometry-awareness property. The second one is to adopt it in a Bayesian network layer. Hierarchical deep Bayesian learning models will be discussed. The feature aggregation will benefit from the nice intra-kernel convolution property.
%We make a theoretic comparison with two classical kernels, the Mat\`ern kernel and the Spectral Mixture~(SM) kernel~\cite{wilson2013gaussian}. The Mat\'ern Kernel is the solution to a linear fractional SPDE~\cite{cressie2015statistics,sherman2011spatial,guttorp2006studies}: $(\kappa^2-\Delta )^{\alpha/2}X(v)=W(v),\quad v\in \mathbb{R}^d$, where $X(v)$ is a Gaussian field function about spatial location $v$, $W(v)$ is the Gaussian white noise, $\alpha$ is a constant, and $\kappa$ is the parameter. This SPDE is mainly used to describe spatial statistics. Our theoretical basis, the reaction diffusion process, is to describe spatial-temporal statistics.
%When the reaction term is zero~(its physical meaning is an equilibrium system in a static state), both SPDEs describe a similar physical progress. 
%Both kernels have the same goal of making the inference fit better with the physical process. The SM kernel is derived from spectral density functions. It has a quite similar formula with the GAC kernel. The main difference is the periodic term. In the GAC kernel, a scale parameter $\phi$ is added to control the periodic property. 

\subsection{Unsupervised Salient Point Selection}
GP regression~(GPR) is widely used in spatial inference known as the ``Kriging'' method~\cite{rasmussen2004gaussian, gao2019gaussian}. Being inspired by the landmarking in face recognition~\cite{wang2014facial}, 
we use a GPR-based unsupervised salient point selection to process the irregular inputs. This method is essentially a manifold learning technique~\cite{kadoury2018manifold,liang2015landmarking,rashmi2019optimal}. Therefore, applying our method also achieves nonlinear data dimensionality reduction. One advantage of GPR is the availability of uncertainty estimation~\cite{rasmussen2004gaussian}. We leverage this advantage and define the saliency score as the variance-based uncertainty~\cite{zidek2003uncertainty}. By iteratively selecting new salient points and adding previously selected points to the prior, we successively collect a set of salient points. A geometry-aware kernel guarantees the salient points are significant to represent the original massive data. This strategy has been successfully applied in~\cite{liang2015landmarking,gao2019gaussian, fan2020morphometric}. 

\begin{figure*}[t]
\centering
\includegraphics[width=\linewidth]{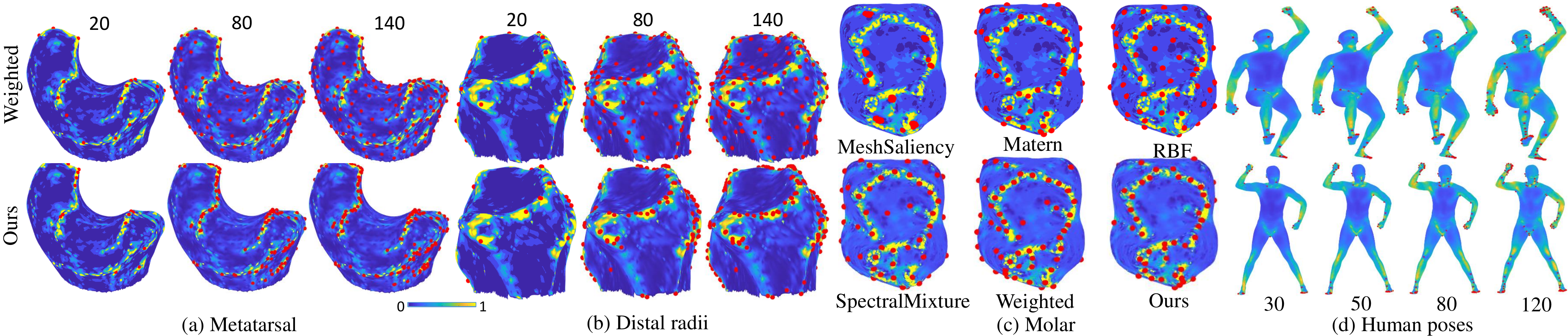}
\caption{Visualization of salient points. All meshes are rendered by the normalized mean curvature. The ROI is estimated to be the high-curvature yellow region. The salient points are marked by the red spheres. (a) and (b) illustrate 20, 80, 140 salient points on metatarsal and distal radii data. The upper row is the results of W-GP~\cite{gao2019gaussianearly}, the bottom row is ours. (c) illustrates 50 salient points selected by comparison methods and ours on a molar model. (d) shows 30-120 salient points on two human poses. The saliency transition is visible.}
\label{fig:saliency_results}
\end{figure*}

Suppose a set of $\kappa$ salient points is denoted as $\tilde{v}=\{\tilde{v}^1,...,\tilde{v}^{\kappa}\}$. Define $\lambda=\sqrt{0.2\pi n},n=[1,...,N_{fre}]$. $\phi$ equals to a $N_{fre}$-length vector by dividing $[0, \frac{\pi}{2}]$ into $N_{fre}$ equal line-spaces. A multi-frequency multi-phase GAC kernel~($MMK$) is defined in a weighted squared form: $MMK=K\times W\times K$~($K$ is symmetric). The weight $W$ is a diagonal matrix with the sum of absolute values of each row in GAC kernel $K$ as the diagonal entries: $W(v)=\sum \left |K(v,\cdot)\right |$. The saliency score $\Sigma_\mathcal{M}$ of $v_i$ during selecting the $({\kappa+1})^{th}$ salient point is defined as:
\begin{equation}\label{eq:23}
\begin{aligned}
\Sigma^{\kappa+1}_\mathcal{M}(v_i) = K(v_i,v_i)-K(v_i,
	\tilde{v}^{\kappa})K_{\tilde{v}^{\kappa},\tilde{v}^{\kappa} }^{-1}K^T(v_i,\tilde{v}^{\kappa})
\end{aligned}
\end{equation}
\begin{equation}\label{eq:24}
K(v_i,\tilde{v}^{\kappa})=\begin{pmatrix}
K(v_i,\tilde{v}^{1})\\ \vdots  
\\ 
K(v_i,\tilde{v}^{\kappa})
\end{pmatrix}_{\kappa\times 1}
\end{equation}
\begin{equation}\label{eq:25}
K_{\tilde{v}^{\kappa},\tilde{v}^{\kappa}}=\begin{pmatrix}
K(\tilde{v}^{1},\tilde{v}^{1}) & \cdots  & K(\tilde{v}^{1},\tilde{v}^{\kappa})\\ 
\vdots  &  & \vdots\\ 
K(\tilde{v}^{\kappa},\tilde{v}^{1}) & \cdots & K(\tilde{v}^{\kappa},\tilde{v}^{\kappa})
\end{pmatrix}_{\kappa\times \kappa}
\end{equation}
Only the point with the highest uncertainty score is selected as the salient point: $\tilde{v}:=argmax_v\Sigma$. The first salient point is the vertex with the maximum variance in $MMK$. From Eq.~\eqref{eq:23}-\eqref{eq:25} we can see that a newly-selected salient point will be added to the prior knowledge and the next saliency score is determined by the previous selections. The whole process follows a GPR framework. %As known, when $n$ points $\left \{ v_1,...,v_n \right \}$ and their corresponding observations are given as the prior knowledge, we can estimate the observation for the $(n+1)^{th}$ new input by evaluating the closeness between the existing points and the new point. Similarly, we can estimate the next salient point after the prior knowledge is updated by the current selection. 
%The following progress is iteratively operated until $\kappa$ salient points $\left \{ \tilde{v}^1,...,\tilde{v}^\kappa\right \}$ are collected and we can see the selection strategy is greedily maximizing the total uncertainty score of the salient point set. This uncertainty estimation method is called the variance-based uncertainty estimation~\cite{zidek2003uncertainty}.%
The algorithm is summarized in Algorithm~\ref{Alg:1}. Figure~\ref{fig:datatype}(b) shows examples of selecting salient points on point clouds. Figure.~\ref{fig:saliency_results} demonstrates salient points on triangle meshes. Further evaluation results are available in Experiments. %This mesh is rendered by normalized mean curvature. High yellow regions stand for high-curvature areas. Figure~\ref{fig:saliency}(b) shows an example of selecting the first three salient points. The mesh is rendered by the normalized saliency scores. We can see this uncertainty map is quite similar to the mean curvature map in Figure~\ref{fig:saliency}(a), which 
%validates that our GAC kernel is geometry-aware. Figure~\ref{fig:saliency}(c) and (d) show the saliency map after selecting 12 and 39 points. We can observe the uncertainty changes after one point is selected. This is reflected as the color fade at this point in figures.
\begin{algorithm}[t]
\centering
\caption{Unsupervised Salient Point  Selection}\label{Alg:1}
\begin{algorithmic}[t]
\Procedure{GPR}{$\mathcal{M},\kappa$}\Comment{Manifold $\mathcal{M}$, $\kappa$ salient points}
\State $N\gets$ KNN or Fast Marching \Comment{calculate $N$ nearest neighbors of each point}
\State $MMK=K\times W\times K\gets$ Kernel construction
\State $\tilde{V}\gets \varnothing$ \Comment{initialize landmarks set $\tilde{V}$ as empty}
\While{$k \leq \kappa$}
\If{k=1}
\State $\Sigma_{\mathcal{M}}(v_i)\gets max(diag(MMK))$
\Else
\State $\Sigma_{\mathcal{M}}(v_i)$  \Comment{calculate the uncertainty score}
\EndIf
\State $\tilde{v}^k \gets argmax\Sigma_{\mathcal{M}}$
\State $k \gets k+1$
\EndWhile
\State \textbf{return} $\tilde{V}=\left \{ \tilde{v}^1,...,\tilde{v}^\kappa \right \}$
\EndProcedure
\end{algorithmic}
\end{algorithm}

\subsection{Hierarchical Bayesian Model on Manifolds}
\begin{figure}[t]
\centering
\includegraphics[width=\linewidth]{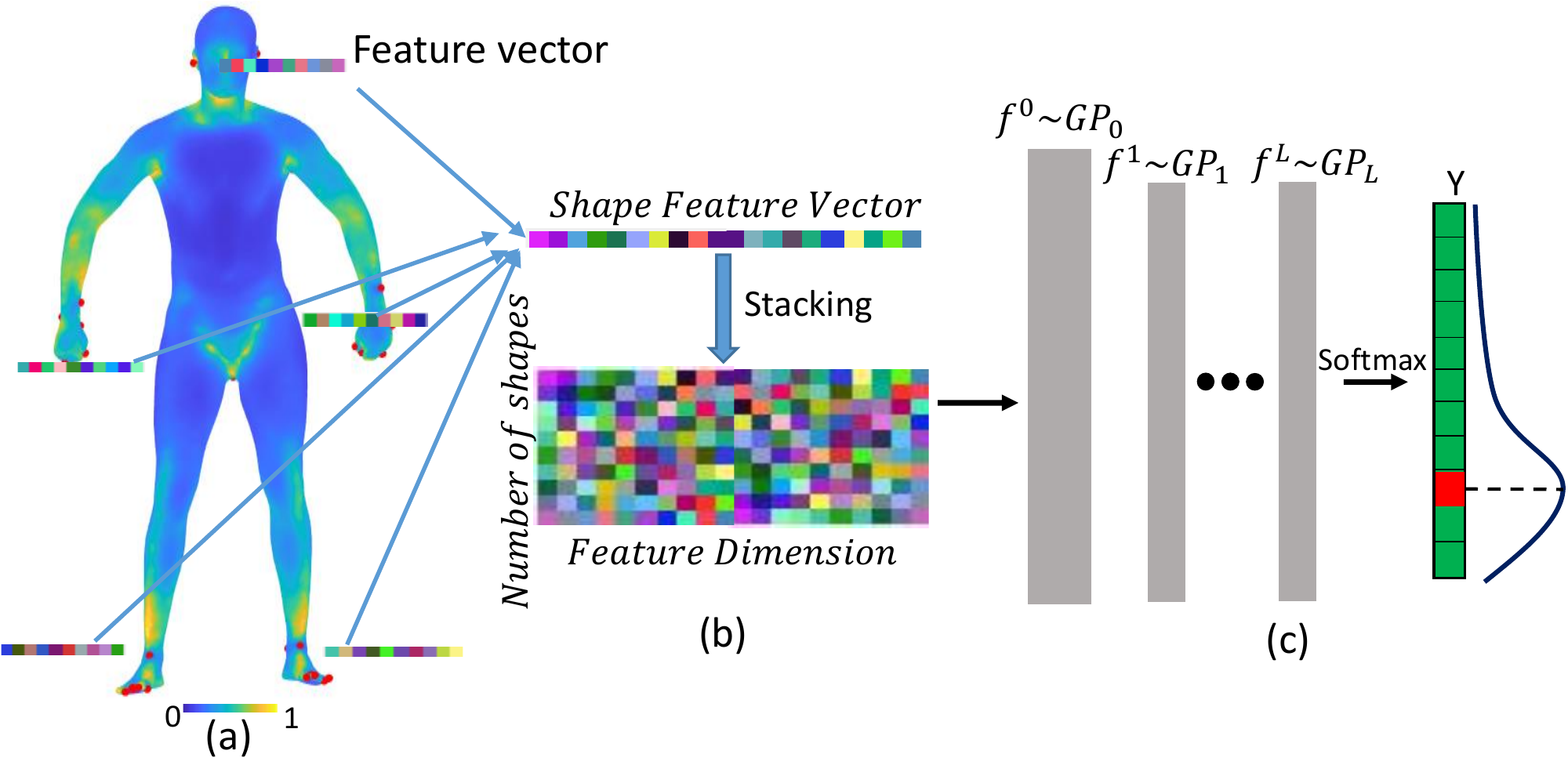}%21
\vspace{-0.5em}
\caption{Pipeline of human pose retrieval. (a) Point-wise feature computation and salient point selection. The mesh is rendered by normalized mean curvatures. (b) Shape feature preparation. (c) Hierarchical Bayesian learning model for feature aggregation and inference. A softmax likelihood function is used at the last layer.}
\label{fig:saliency_features}
\end{figure}

We define a GAC-GP layer with the GAC kernel and follow the framework of DGPs to construct a hierarchical Bayesian learning model by stacking up multiple GP layers. Thanks to the intra-kernel convolution property, the GAC-GP layer has a good feature aggregation ability. %Noting that the input of a DCGP is the collection of features, therefore, we actually compute the covariance of features and the convolutional operation in the kernel computation helps to find the patterns of features. The convolutional filter is implicitly embedded in the kernel definition and it is optimized with the kernel parameter tuning.
In a pure hierarchical Bayesian learning model on manifolds, a computational pipeline is shown in Figure.~\ref{fig:saliency_features}. The first step is to process the input. Assume $\kappa$ salient points are selected by Algorithm~\ref{Alg:1}. The feature on each salient point $f_{\tilde{v}}$ is a vector of length $l$. For shape $i$, we link all features of salient points in the order of their selections: $f_i = \{f_{\tilde{v}^1}...f_{\tilde{v}^{\kappa}}\}$. Noting that all shapes here belong to the same dataset and all salient points are selected with the same parameter setting in Algorithm~\ref{Alg:1}. Otherwise, point-to-point registration is needed to concatenate features. Suppose $H$ shapes are used, the input $X$ is an $H\times (\kappa\times l)$ matrix. We compose a sequence of layers that map the input $x_i$ to its label $y_i$ in a hierarchical Bayesian model for classifications:
\begin{equation}\label{eq:dgp}
\vspace{-0.5em}
\underset{1\times (\kappa\times l)}{\underbrace{x_i=f^0}}\overset{\mathcal{GP}_0}{\rightarrow}\underset{1\times S^1}{\underbrace{f^1}}\rightarrow \cdots \overset{\mathcal{GP}_{L-1}}{\rightarrow}\underset{1\times C}{\underbrace{f^L}}\overset{softmax}{\rightarrow} \underset{C_i}{\underbrace{y_i}}
\end{equation}
The output of hidden layer $l$ is a vector of the size $1 \times S^l$, where $S^l$ is the layer size. This is similar to the relationship of the input channel and output channel in a NN. When the batch processing is applied, the output of each hidden layer has dimension $B\times S^l$, $B$ is the batch size. A final layer is appended with a softmax multi-class likelihood. The output vector has the dimension $1\times C$, where $C$ is the number of classes. Each entry stands for the probability belonging to a certain class. Arbitrary numbers of GP layers can be added as hidden layers. We use the doubly stochastic variational inference approach to estimate the posterior~\cite{salimbeni2017doubly}. The optimization process is to maximize the ELBO in Eq.~\eqref{eq:elbo}. The K-means method is used to choose inducing points.

Because the input of the Bayesian model is the point-wise features on manifolds, and NNs are strong in feature learning, we are inspired to further explore the potential of NN+Bayesian methods. Such a mixed model can take advantage of both the feature learning ability of NNs and the feature aggregation ability of Bayesian models. We generally follow the pipeline in deep kernel learning~\cite{wilson2016deep}. The input format is determined by the NN part. The output of NNs is the shape feature. The feature is then fed into a Bayesian model, and the following processing is the same as the pure Bayesian method. The negative marginal log-likelihood~(MLL) is used as the loss function.

\section{Experiments}
We evaluate our methods with three experiments. Our method is noted as GAC-GP. Applications are implemented in Pytorch and GPytorch with GPU acceleration~\cite{gardner2018gpytorch}.

\begin{figure*}[t]
\centering
\includegraphics[width=\linewidth]{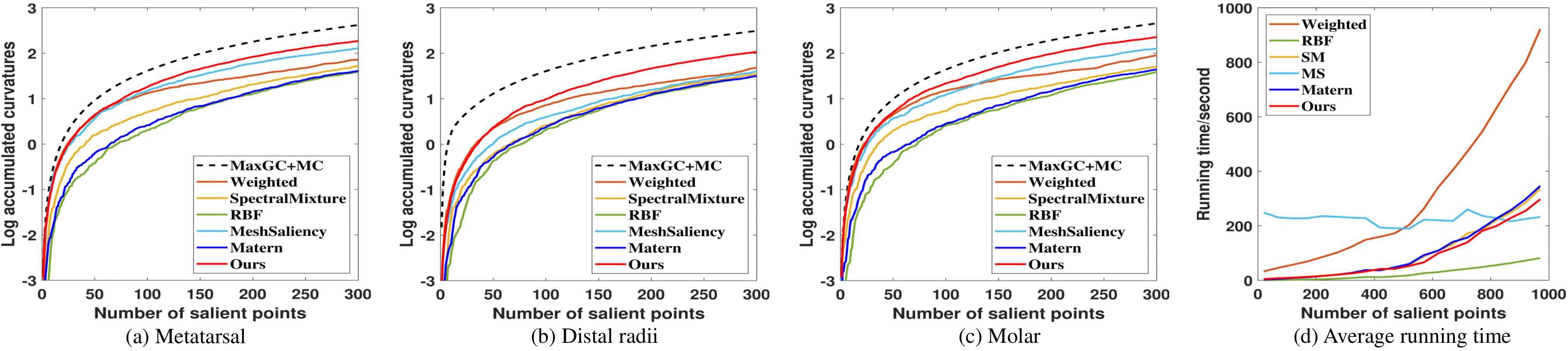}
\vspace{-1.2em}
\caption{(a)-(c) The log accumulated AC curve of each dataset. (d) Average running time of selecting one salient point.}
\label{fig:curves}
\end{figure*}

% \textbf{Comparison methods}: (1) RBF kernel GP~(RBF-GP)~\cite{rasmussen2004gaussian}; (2) Spectral mixture kernel GP~(SMK-GP) in~\cite{wilson2014covariance}. 10 mixtures are used;
% %We also test some other successful kernels such as the Mat\'ern kernel family~\cite{stein1991kernel,stein2012interpolation}. But the performances are not comparable, so we neither demonstrate them here; 
% (3) Mat\'ern kernel GP~(Matern-GP)~\cite{stein1991kernel};
% (4) Mesh saliency~(MS)~\cite{lee2005mesh}. This is a highly cited classical method in saliency detection on meshes; (5) Weighted GP~(W-GP) in~\cite{gao2019gaussianearly}. W-GP is the current state-of-the-art GP method on manifolds. We directly use their parameter settings; (6) Graph convolutional GP~(GCGP) in~\cite{walker2019graph}.
\subsection{Unsupervised Salient Point Selection}\label{sec:trianglemeshresults}
In the first experiment, the task is to select salient points on manifolds. The purpose is to evaluate the geometry-awareness of the GAC-GP and its stability in continuous regressions. Additionally, we compare the computational efficiency by recording the average running time. When defining $MMK$, we use the fast marching algorithm to compute geodesic distances and only select $k=200$ nearest neighboring points for each vertex. $N_{fre} $ is set as 5.

Three datasets are used~\cite{boyer2011algorithms}: (i) ``Mandibular molars'', or ``molar'', contains 116 teeth shapes; (ii) ``First metatarsals'', or ``metatarsal'', contains 57 shapes; and (iii) Distal radii contains 45 shapes. All shapes are triangle meshes with around 5000 vertices. Examples of each dataset are illustrated in Figure~\ref{fig:saliency_results}(a)-(c). The ROIs of such data are usually the marginal ridges, teeth crowns, and outline contours where the geometric features are rich~\cite{boyer2011algorithms,couette20103d}. Therefore, the meshes in Figure.~\ref{fig:saliency_results}(a)-(c), are rendered by the normalized mean curvature as the ground-truth~\cite{guennebaud2007algebraic,guennebaud2008dynamic}. The salient points are expected to evenly distribute in yellow regions.  

Comparison methods include: (1) RBF kernel GP~(RBF, RBF-GP)~\cite{rasmussen2004gaussian}. RBF is a classical choice; (2) spectral mixture kernel GP~(SM, SpectralMixture, SMK-GP)~\cite{wilson2014covariance}. SMK-GP had good performances in many vision tasks. 10 mixtures are used; (3) Mat\'ern $3/2$ kernel GP~(Mat\'ern, Mat\'ern-GP)~\cite{stein1991kernel}. Mat\'ern $1/2$ and $5/2$ are excluded because of worse results;
(4) mesh saliency~(MeshSaliency, MS)~\cite{lee2005mesh}. It is a classical saliency detection method; (5) Weighted GP~(Weighted, W-GP) in~\cite{gao2019gaussianearly}. W-GP is a state-of-the-art GP method on manifolds. The same parameter settings are used. %(6) Graph convolutional GP~(GCGP) in~\cite{walker2019graph}. Since.

Figure~\ref{fig:saliency_results}(a)-(c) demonstrate salient point selection results. In (a) and (b), we focus on comparing ours with the W-GP because it is the only GP kernel method that stresses the geometry-awareness, and it outperforms all other comparison methods. We show 20, 80, and 140 salient points. When selecting a small number of salient points, both methods present reasonable results. But the accuracy of W-GP gradually drops with the iterations increasing while GAC-GP shows a much better stability of geometry-awareness. For other comparison methods, we give examples of 50 salient points on one Molar shape in Figure~\ref{fig:saliency_results}(c). Figure.~\ref{fig:datatype}(b) shows some examples of salient points on point clouds. %It shows that salient points recognized by GAC-GP fit better with expectations as most of them are clustered in significant regions. And GAC-GP shows a better consistency in consecutively making reasonable inferences. For example, the results of selecting 140 salient points show that W-GP fails to give correct posterior inferences. The same phenomenon also happens on other comparison methods.

Numerically, we define an Accumulated Curvature~(AC) value to measure the selection performance: $AC_N=log\sum _{k=1}^\kappa (\left | GC_k \right |+\left | MC_k \right |)$, where $GC$ and $MC$ are normalized Gaussian curvature and mean curvature, and $\kappa$ is the total number of salient points. Drawing the AC values with the increased number of salient points forms an AC curve. Within a dataset, we compute the log average of AC values to plot an average AC curve. This curve reflects the geometry-awareness of different selection methods. When only selecting points with the largest accumulated curvatures at each iteration and draw the AC curve with these maximum values, we get a MaxGC+MC AC curve. This curve stands for the upper bound of saliency selections. The higher and closer to this MaxGC+MC curve an AC curve is, the better the geometry-awareness ability a corresponding method has. The results are plotted in Figure~\ref{fig:curves}(a)-(c). The MaxGC+MC AC curve is drawn in dashes. Generally, AC curves of GAC-GP are above all other methods. Both the empirical visualizations and the numerical measurements verify that the GAC-GP is capable of making geometry-aware inference on manifolds. More importantly, its geometry-awareness is still consistent and stable after continuous regressions.

The computational efficiency is measured by averaging the running time of selecting one salient point, as shown in Figure~\ref{fig:curves}(d). GPR-based methods gradually slow down due to the increased prior knowledge as shown in Eq.~\eqref{eq:23}-\eqref{eq:25}. The GAC-GP enjoys strong computational efficiency considering its superior performance because usually only a small number of salient points are needed.

\subsection{Human Pose Retrieval}
In the second experiment, the task is to classify different human poses modeled by triangle meshes. The first purpose is to further evaluate the salient point selections by fixing the Bayesian learning architecture. The second purpose is to fix the inputs and evaluate different hierarchical Bayesian learning architectures. The pipeline in Figure.~\ref{fig:saliency_features} is used. We choose the scaled-invariant wave kernel signature~(SIWKS)~\cite{li2018scale,aubry2011wave} as the feature. When computing the SIWKS, 30 smallest eigenvalues are used. The other parameter settings are the same as those in~\cite{aubry2011wave}. Features of 50, 100, and 250 salient points are used. In customizing the Bayesian model, we use the multitask variational strategy and the softmax likelihood in GPytorch. The number of inducing points is 50. We use Adam as the optimizer with an initial learning rate of 0.001. After the first 200 epochs, the learning rate changes to $10^{-4}$. We trained for 2000 epochs. The cost function is the variational ELBO mentioned in Sec~\ref{sec:HBM}. The comparison methods are the same as the prior experiment.

The SHREC14 non-rigid 3D human model is used~\cite{pickup2014shrec}. It contains 400 triangle meshes of 40 human subjects with 10 poses. Each mesh has about 15000 vertices. The dataset is randomly split into: 90\% for training, 5\% for validation, and 5\% for testing.
%We tested both the Scale Invariant Heat Kernel Signature~(SIHKS)~\cite{bronstein2010scale} and the SIWKS as the pointwise features. But the results with SIHKS are generally worse ($< 0.75$) than those of SIWKS, so we only report the results with SIWKS. 
\begin{table}[t]
\centering
\resizebox{\textwidth}{!}{%
\begin{tabular}{c|cccccc}
\hline
SIWKS & RBF-GP & W-GP & Matern-GP & SMK-GP & MS & GAC-GP \\ \hline
50 & 0.850 & 0.898 & 0.885 & 0.898 & 0.866 & \textbf{0.915} \\ \hline
100 & 0.859 & 0.901 & 0.886 & 0.908 & 0.872 & \textbf{0.921} \\ \hline
250 & 0.862 & 0.905 & 0.890 & 0.912 & 0.899 & \textbf{0.925} \\ \hline
\end{tabular}%
}
\caption{Results of human pose retrieval with Bayesian models defined by different kernels and numbers of salient points.}
\label{tab:class1}
\end{table}
For the first purpose, we fix the Bayesian model to be a one-layer GAC-GP and feed in features from different methods. Table~\ref{tab:class1} shows the results. Classification with the features of all points has an accuracy of 0.910. Taking this value as a reference, we can draw conclusions that (1) our strategy of selecting salient points works for distinguishing different shapes. When enough salient points are selected, it is possible to use a small subset to represent the original data; (2) the geometry-aware selection of GAC-GP is more distinguishable than other comparison methods. 
\begin{table}[t]
\centering
\resizebox{\textwidth}{!}{
\capbtabbox{
\begin{threeparttable}
\begin{tabular}{ccc}
\hline
\textbf{Method} & & \textbf{Accuracy} \\ \hline
GCGP* & & $91.2\%$ \\
1RBF & & $91.1\%$ \\
1GAC & & $92.5\%$ \\
1GAC(10)+1RBF & & $92.7\%$ \\
1GAC(10)+1GAC & & $\textbf{93.4\%}$ \\ \hline
\end{tabular}
\begin{tablenotes}
\small
\item *(Using self-reproduced code.)
\end{tablenotes}
\end{threeparttable}
}
{
\caption{Human pose retrieval with different Bayesian learning architectures.}
\label{tab:class2}
}
}
\end{table}
\begin{table}[t]
\resizebox{\textwidth}{!}{
\capbtabbox{
\begin{tabular}{cc}
\hline
\textbf{Method} & \textbf{Error rates} \\ \hline
PCNN & $86.1\%$ \\
% PointNet & $89.2\%$ \\
PointNet++ & $90.7\%$ \\
PointNet++ +Normal & $91.9\%$ \\ \hline
PCNN+1GAC & $\textbf{87.2\%}$ \\
% PointNet++ +1RBF & $91.6\%$ \\
PointNet++ +1GAC & $\textbf{91.8\%}$ \\
PointNet++ +Normal+1GAC & $92.1\%$\\
PointNet++ +Normal+2GAC & $92.8\%$ \\ 
PointNet++ +Normal+3GAC & $\textbf{93.1\%}$ \\\hline
\end{tabular}
}{
 \caption{Multi-class classifications on ModelNet40.}
 \label{tab:modelnet40}
}
}
\end{table}
% \begin{table}[t]
% \centering
% \resizebox{0.5\textwidth}{!}{%
% \begin{tabular}{cc}
% \hline
% Methods & Accuracy \\ \hline
% GCGP* & 0.912 \\
% 1RBF & 0.911 \\
% 1GAC & 0.925 \\
% 1GAC(500)+1RBF & 0.927 \\
% 1GAC(10)+1RBF & 0.927 \\
% 1GAC(10)+1GAC & 0.934 \\ \hline
% \end{tabular}%
% }
% \caption{Classifications with different Bayesian learning architectures. Features of 250 salient points are used in our methods. 50 inducing points are used in all methods. The output size of each layer is in brackets.}
% \label{tab:class2}
% \end{table}
For the second purpose, we fix the inputs to be GAC-GP salient features and evaluate different Bayesian learning architectures. Here we use GCGP~\cite{walker2019graph} as a comparison method. The results are shown in Table.~\ref{tab:class2}. Noting that the code of GCGP is not available and we use our implementations, so we put a star mark on GCGP's result. We can see that the accuracy is generally increased after adding a GAC layer, supporting a strong feature aggregation property. Meanwhile, a hierarchical concatenation of GAC layers shows a better accuracy than the single layer structure. %More experimental results are available in Supplementary. %Due to the computer memory limitation, we did not test models with more GAC layers and more compositions. We also test the relationship between the size of hidden layers and the final performance by gradually changing the size from 500 to 10. But we observe no obvious differences.

% \usepackage{graphicx}

% \usepackage{graphicx}
% \begin{table}[t]
% \centering
% \resizebox{0.65\textwidth}{!}{%
% \begin{tabular}{cc}
% \hline
% Methods & Accuracy \\ \hline
% PCNN & 0.861 \\
% PointNet & 0.892 \\
% PointNet++ & 0.907 \\
% PointNet++ +Normal & 0.919 \\ \hline
% PCNN+GAC & 0.872 \\
% PointNet++ +GAC & 0.918 \\
% PointNet++ +Normal+GAC & 0.921 \\ \hline
% \end{tabular}%
% }
% \caption{Multi-class classifications on ModelNet40.}
% \label{tab:modelnet40}%\label{tab:my-table}
% \end{table}
%In this paper, we focus on using triangle meshes, but we can make a reasonable infer that our methods are also feasible to other types of manifold-valued data. In supplementary material, we demonstrate a set of salient point selection results on the McGill 3D shape benchmark~\cite{siddiqi2008retrieving}. We remove the edge connections of each shape and assume the data is a point cloud. Our future work will focuse on designing an end-to-end deep Bayesian network. An interesting topic is to fuse the feature extraction into a deep GP model like the feature extraction in neural networks. A fully integrated deep Bayesian model or a deep kernel learning model~\cite{wilson2016deep} may be even more expressive and applicable for manifold-valued data.
\subsection{Point Cloud Classification}
In the third experiment, the task is to classify different point cloud models. Our purpose is to demonstrate the work of integrating NNs with Bayesian learning. Here, we use the hierarchical feature learning architecture in PointNet++~\cite{qi2017pointnet++} to learn the pointwise feature. We perform multi-class classification on ModelNet40 which contains 12311 3D CAD models of 40 categories. Each point cloud has 10000 points. We use 9843 models for training and 2468 models for testing. In the feature aggregation part, we use one single GAC-GP layer~(ten mixtures). 64 inducing points are used. The optimizer is Adam and the initial learning rate is 0.04. The comparison methods include PointNet++, PointNet++ with normal information, and the Pointwise Convolutional NNs~(PCNN)~\cite{hua2018pointwise}. Table~\ref{tab:modelnet40} shows that (1) the mechanism of NN+Bayesian can be jointly trained for tasks on manifolds; (2) models with Bayesian aggregation layers generally outperforms the classical multiple fully connected layers in our tests. We notice that the performance gain of using single GAC layer shrinks after adding normal information. Our hypothesis is that the features become more complicated, and the inference capability of single GAC layer is not powerful enough to well aggregate the new features. By adding 2\&3 GAC layers, the improvements increase to 0.9\% and 1.2\%, respectively. The overall results show that architectures with GAC layers universally perform better than their original versions, which proves that such a co-design benefits the performance. A reasonable outlook is to investigate more effective architectures that integrate both methods for end-to-end tasks on manifolds.%It is also clear to see that the input feature is crucial for a Bayesian learning method on manifolds. %So far, learning features might be a weak point of Bayesian learning methods. 
%The results are remarkable since they may inspire a co-learning mechanism to combine strong feature learning (NN) and efficient inference (Bayesian) abilities.  
\section{Conclusion}
In this work, we propose the GAC kernel that carries properties of geometry-awareness and intra-kernel convolution. Our methods show strong feature aggregation capability in various tasks on manifolds. We hope our work may inspire future Bayesian and NN+Bayesian studies on manifolds.

\noindent\textbf{Acknowledgements} This work was funded by grants R01EY032125 and R21AG065942.
%\subsection{References}

\section{Supplementary Materials}
\subsection{Theorem1 and Proof}
\begin{customthm}{1}%\label{theorem:1}
A real-valued function $T(v,t)$ on $\mathbb{R}^d$ is a spatial-temporal kernel function if it is a linear/non-linear diffusion process: $\frac{\partial T}{\partial t}=\alpha \Delta T + P(t)\delta(v)$, where $\alpha$ is a positive constant, $P(t)$ is a periodic function, $\delta(v)$ is the Dirac delta function, and $\Delta$ is the Laplace operator.
\end{customthm}

\begin{proof}
As known, there exists a corresponding Green's function $G(v, v',t, t')$ for a parabolic partial differential equation so that the diffusion process expressed by this parabolic partial differential equation has the form~\cite{ehrlich1980surface,strauss2013partielle}:
\begin{equation}\label{eq:diffusion}
T(v, v',t)=\int_{0}^{t}G(v, v',t-s)P(s)ds
\end{equation}
In the main submission we have proved that the analytical solution to the following equation is PSD:
\begin{equation}\label{eq:analytical}
T=\int_{0}^{t}G(v, v',t-s)cos\omega(t-s)ds
\end{equation}

Since the spatial variable mainly exists in the Green's function and the Green's function is spatial stationary in a $\mathcal{R}^d$ diffusion process, the primary task is to prove any choices of periodic function $P(t)$ can lead to the same conclusion. Because Eq.~\eqref{eq:analytical} has been prove to be PSD, we can draw the same conclusion if Eq.~\eqref{eq:diffusion} has a similar form with Eq.~\eqref{eq:analytical}. Therefore, the main idea is using cosine function to generalize a periodic function $P:\mathbb{R}\rightarrow \mathbb{R}$. As known, a periodic function can be estimated with the Fourier series expansion: 
\begin{equation}\label{eq:fourier}
P(t) = \frac{1}{2}a_0+\sum_{n=1}^{\infty}\left [a_ncos(nt)+b_nsin(nt) \right ]
\end{equation}
where $a,b$ are arbitrary real numbers. Assuming there exists a cosine function $x_ncos(nt+y_n)$ that is equal to $a_ncos(nt)+b_nsin(nt)$, $n$ is a constant. By using the trigonometric sum formulae, we get: $x_n = \pm \sqrt{a_n^2+b_n^2}$ and $y=arctan(-\sqrt{\frac{b_n}{a_n}})$. Eq.~\eqref{eq:fourier} is then transformed to:
\begin{equation}\label{eq:pt}
P(t) = \frac{1}{2}a_0 \pm\sum_{n=1}^{\infty} \sqrt{a_n^2+b_n^2}cos\left [nt+ arctan(-\sqrt{\frac{b_n}{a_n}})\right]   
\end{equation}
Eq.~\eqref{eq:pt} shows that any periodic functions can be approximated by the linear combination of cosine functions. By substituting Eq.~\eqref{eq:pt} into Eq.~\eqref{eq:diffusion}, we see that the diffusion process is estimated to be the integral of Green's function times the combination of cosine functions. Applying the PSD summation identity, we can draw the conclusion that the solution to Eq.~\eqref{eq:diffusion} is also PSD. Therefore, it is a valid kernel function. 
\end{proof}
\begin{figure}[t]
\centering
\includegraphics[width=\linewidth]{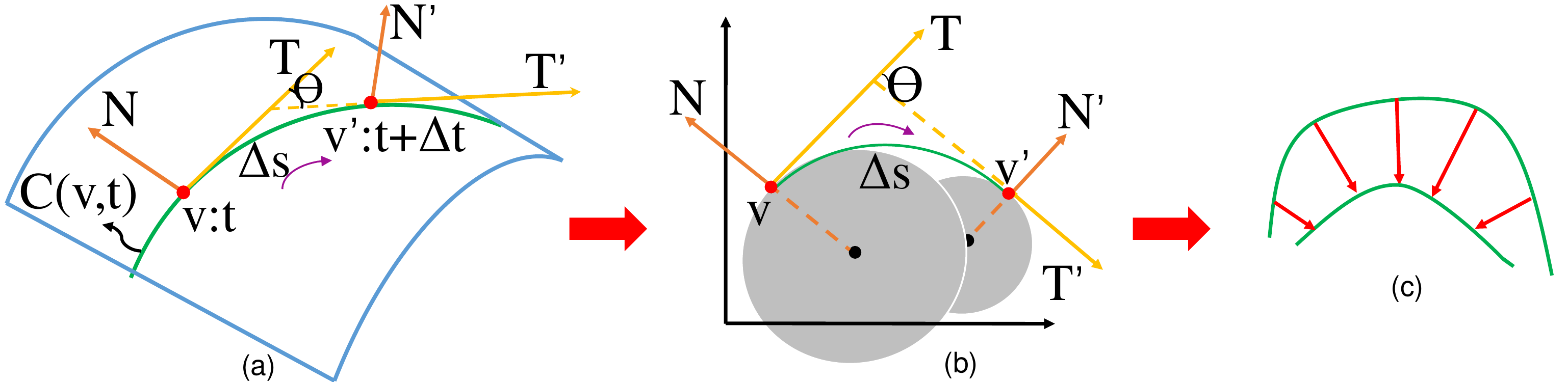}
\caption{Sketch plots of (a) a planar curve example; (b) the curve in a 2D plane. The shaded circles are inscribed circles on $v$ and $v'$; (c) the curve moves from top to down by curvature flow.}
\label{fig:curvatureflow}
\end{figure}
\subsection{Lemma 1 and Proof}
\begin{customlemma}{1}%\label{lemma:1}
The GAC Kernel embeds the mean curvature flow in $\mathbb{R}^3$, which enables it to be geometry-aware.
\end{customlemma}

\begin{proof}
In differential geometry, a curvature flow numerically links intrinsic geometric features and extrinsic flows together~\cite{kichenassamy1995gradient}. %The general solution of the reaction diffusion process essentially describes a diffusion problem. 
We take the proof on a planar curve by assuming manifold $\mathcal{M}$ is a two dimensional manifold in $\mathbb{R}^3$ as an example for convenience. In this case, the GAC kernel is actually equivalent to a curve-shortening flow which can be considered as a one dimension mean curvature flow~\cite{altschuler1993shortening}. Figure~\ref{fig:curvatureflow} shows sketch plots of the symbols used in this proof. Suppose $v$ is a point on the manifold. $C(v)$ is the intersection between the manifold and the normal plane on $v$. As known, $C(v)$ is a 1-dimensional smooth curve. Assume one point moves along $C$ from $v$ to $v'$. Let $\Delta s$ be the arc length of this movement and $\theta$ be the rotation angle of the tangent vector, then we can define the following concepts: \newline(i) the velocity vector at $v$ is $\frac{dC}{dv}$; \newline(ii) the velocity is the magnitude of the velocity vector, which is $\left | \frac{dC}{dv} \right |=\frac{ds}{dv}$; \newline(iii) the unit tangent vector $T=\frac{dC}{ds}/\left|\frac{dC}{ds}\right|$ and the unit normal vector $N=\mathcal{R}T$. $\mathcal{R}$ is a $\pi/2$ rotation matrix; \newline(iv) the curvature $\kappa$, which measures how fast the unit tangent vector rotates relative to the arc length: $\kappa = \underset{\Delta s\rightarrow 0}{lim}\left|\frac{\Delta \theta}{\Delta s}\right|$. And we can further get $\frac{dT}{ds}=(-sin\theta, cos\theta)\frac{d\theta}{ds}=\kappa N$ and similarly $\frac{dN}{ds}=-\kappa T$. 
\begin{figure}[t]
\centering
\includegraphics[width=\textwidth]{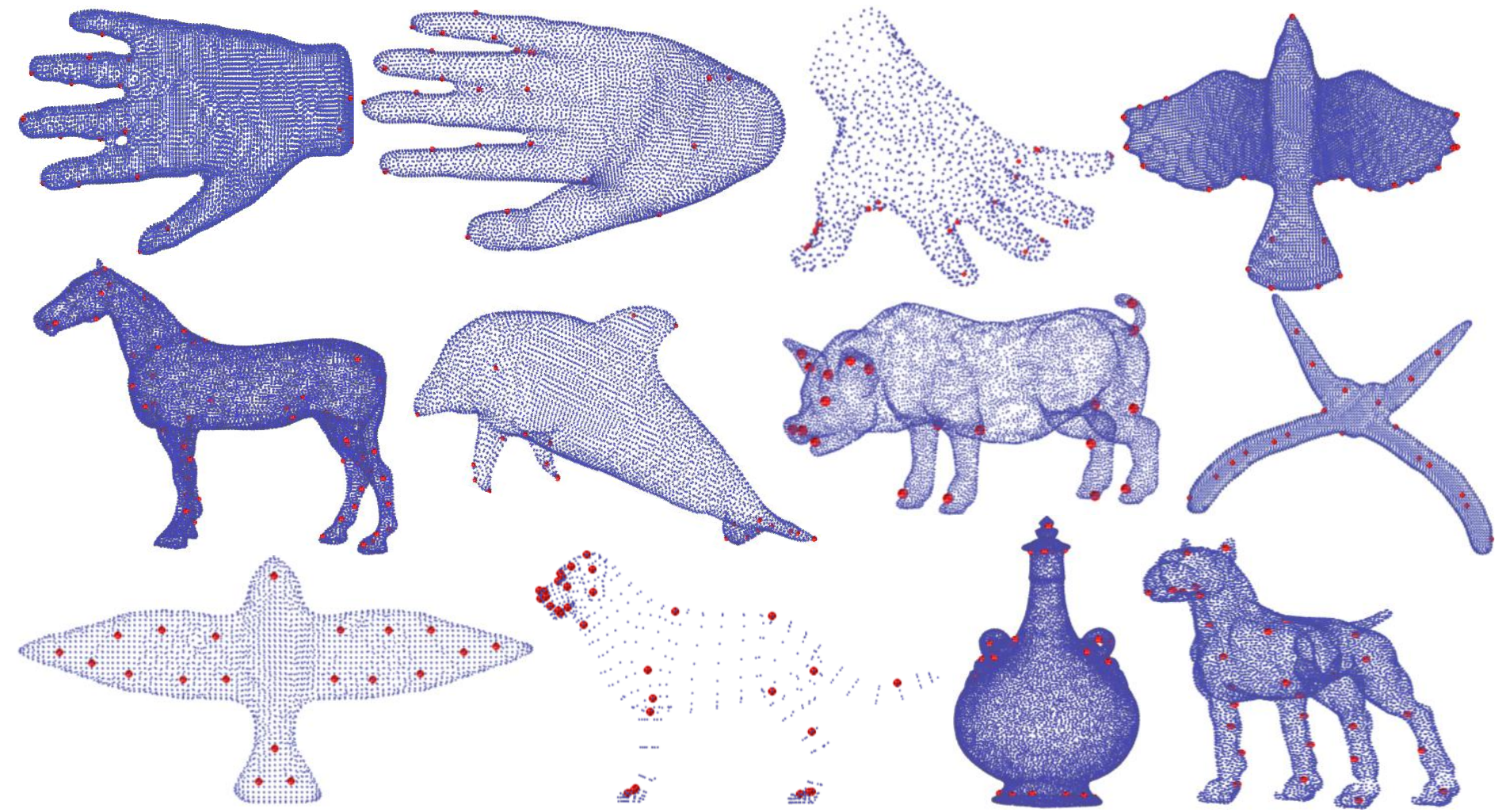}
\caption{Illustration of saliency selection on McGill 3D shape benchmark~\cite{siddiqi2008retrieving}. The salient points are marked by red spheres.}
\label{fig:pointcloud}
\end{figure}

Assume all points on the curve start to move along their normal directions at a velocity of $\kappa(v)$ during time $t$, we have the curvature flow: $\frac{dC}{dt}=\kappa N$. With the equation (iii) and (iv), we write the curvature flow as: $\frac{dC}{dt}=\frac{d}{ds}\frac{dC}{ds}$, which is clearly a diffusion process with a zero reaction function.  This equivalence proves that the GAC kernel can theoretically reflect the geometric features. From the perspective of physical meanings, $G(v,v')$ means how much curvature changes from point $v$ to its neighborhood $v'$ during a period of time. Therefore, the physical meanings also support that the GAC kernel embeds the geometric information of the manifolds. Lemma 1 is proved.
\end{proof}

\subsection{Lemma 2 and Proof}
\begin{customlemma}{2}%\label{lemma:2}
The GAC Kernel embeds a convolution filtering within the kernel structure, called intra-kernel convolution.
\end{customlemma}
\begin{proof}
According to the reaction diffusion theory~\cite{kuttler2011reaction,strauss2013partielle}, Eq.~\eqref{eq:diffusion} can be expressed as: 
\begin{equation}\label{eq:2}
T=\int_0^t\int_{-\infty}^\infty G(v-v',t-s)P(v',s)dv'ds
\end{equation} 
If integrating along the temporal variable, then the result has the form $\int P(v')G(v,v')dv'$, which matches with the definition of a convolutional filtering $\int f(v')h_t(v,v')dv'=(f_0*h_t)(v)$. Our kernel derivation also indicates the existence of a convolution on manifolds. Reminding that we estimate the integral in Eq.~\eqref{eq:diffusion} as the summation of a sine Fourier transform and a cosine Fourier transform~(Eq.14 in the main submission). Each term implements the transform from time domain to frequency domain. According to the Convolution Theorem, we can draw the same conclusion. The similar theory has also been applied in geometric deep learning to realize the convolution on manifolds~\cite{bronstein2017geometric}. %The term $\int f(v')h_t(v,v')dv'$ is the general solution to a diffusion process. The $h_t(v,v')$ is the well-known heat kernel in the thermal diffusion scenario. 
%From the perspective of physical meanings, according to the Lemma1, $G(v,v')$ means how much curvature changes from point $v$ to $v'$ at a short period of time. %In a local domain, the $h_t(v,v')$ is shift-invariant, therefore, we have $h_t(v,v')=h_t(v-v')$. Bringing this equivalency back to the above equation, the term $\int f(v')h_t(v,v')dv'$ representation a convolution filtering $\int_{\mathcal{X}} f(v')h_t(v,v')dv'=(f_0*h_t)(v)$. 
%The term $\int_{0}^{t}e^{-(t-s)\Delta}F(s)ds$ only describes the properties of the reaction part. So the GAC kernel can be regarded as the summation of a convolution filtering and a constant setting within a certain period of time.
Lemma 2 is proved.
\end{proof}

\begin{figure}[t]
\centering
\includegraphics[width=\textwidth]{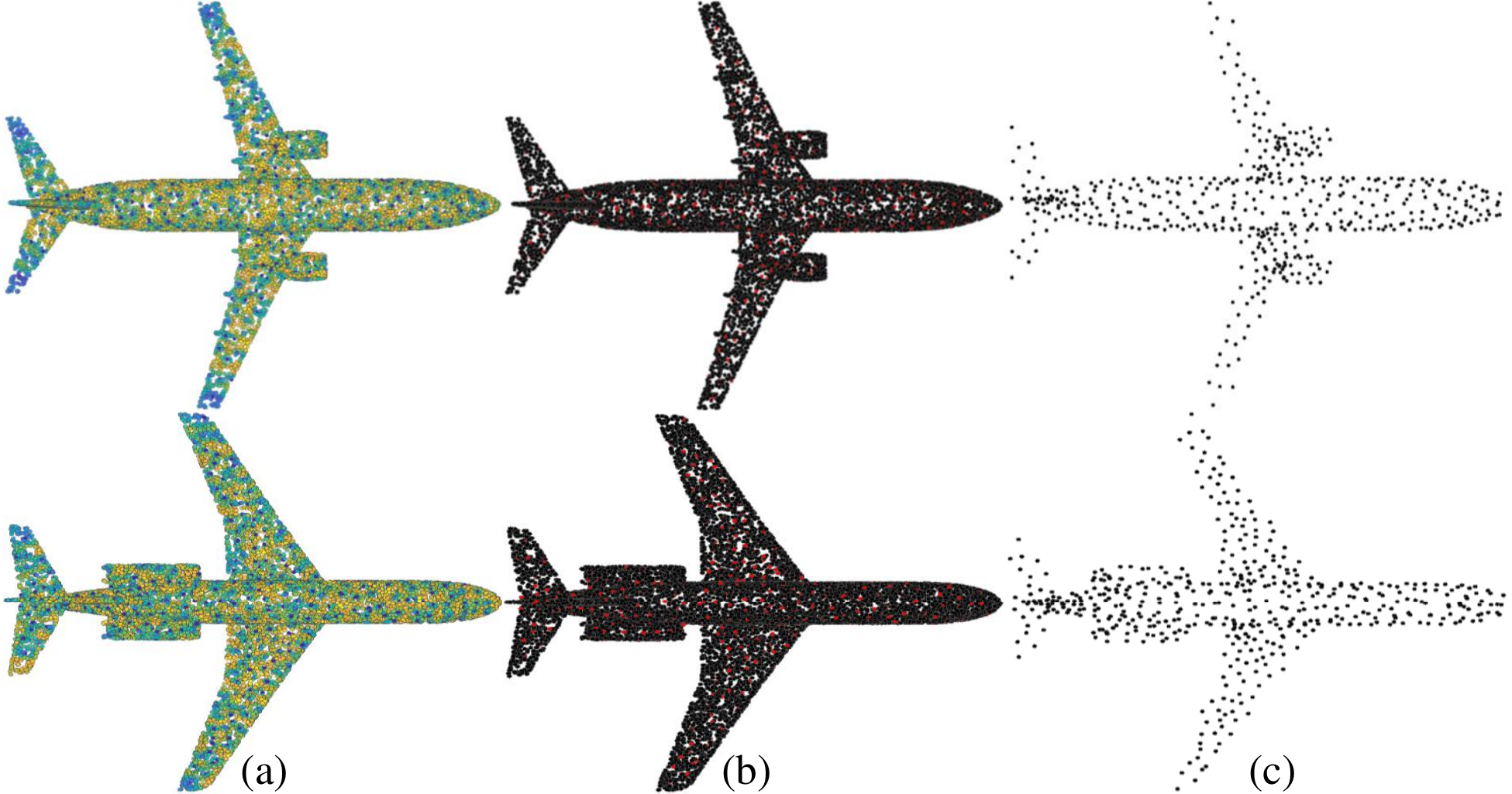}
\caption{Illustration of saliency selection on Modelnet40~\cite{wu20153d}. (a)Demonstrations of saliency maps after selecting 200 salient points. (b)Demonstrations of 200 salient points. The salient points are marked by red spheres. (c) The selected 200 salient points out of 10000 vertices can generally represent the original shape.}
\label{fig:pointcloudmodelnet}
\end{figure}

\subsection{Unsupervised Salient Point Selection}
We provide more salient point selection results on point clouds here. Figure.~\ref{fig:pointcloud} shows twelve examples of salient point selection on McGill 3D shape benchmark~\cite{siddiqi2008retrieving}. The shapes have different sampling densities. The original data type is the triangle mesh. We remove their edge connections and only use the vertex coordinates as inputs. The results show that the GAC-GP can learn the geometric property of the inputs in the prior and the selected salient points are representative in distinguishing the shapes.

Figure.~\ref{fig:pointcloudmodelnet} illustrates two examples of salient point selection on Modelnet40 dataset. Figure.~\ref{fig:pointcloudmodelnet} (a) demonstrates the saliency maps after selecting 200 salient points. Figure.~\ref{fig:pointcloudmodelnet} (b) shows salient points on the original point clouds. Figure.~\ref{fig:pointcloudmodelnet} (c) is 200 salient points. We can see that 200 salient points can generally depict the original shapes. Noting that we did not use salient point selection algorithm in the point cloud classifications in the main submission. This is mainly because the comparison methods used the whole data as inputs, for fairness and convenience, we also use 10000 vertices as inputs in the experiments.

% \section{Human Pose Retrieval}
% In the main submission, we mainly provide experimental results of using hierarchical Bayesian models. Here, we demonstrate two more results with single Bayesian layer in Table.~\ref{tab:class1}. The purpose of adding these two experiments is to (1) test the feature aggregation capability of using single Bayesian layer with different kernels; (2) compare the performance of using one and more than one Bayesian layers. We can see that the GAC-GP layer has a stronger aggregation capability comparing with the classical RBF kernel. Using more than one Bayesian layers help to obtain a better result. Due to the limited computational resources, we did not test deeper architecture. In our hierarchical Bayesian models, we set the output size of the first layer as the number of classes. We also tested different layer sizes, but no obvious differences are observed.

{\small
\bibliographystyle{ieee_fullname}
\bibliography{egbib}
}

\end{document}